\documentclass[journal]{IEEEtran}
\usepackage{algorithm}
\usepackage{algpseudocode}
\usepackage{amsmath,amssymb,amsthm}
\usepackage{array}
\usepackage{bm}
\usepackage{booktabs}
\usepackage{caption}
\usepackage{cite}
\usepackage{empheq}
\usepackage{hyperref}
\usepackage{xcolor}

\hypersetup{
  colorlinks=true,
  linkcolor=black,
  citecolor=black,
  urlcolor=blue
}

\definecolor{codegreen}{rgb}{0.0,0.5,0.0}
\definecolor{codegray}{rgb}{0.4,0.4,0.4}
\definecolor{codepurple}{rgb}{0.5,0,0.5}

\newcommand{\real}{\mathbb{R}}
\newcommand{\Exp}{\text{Exp}}
\newcommand{\Log}{\text{Log}}
\newcommand{\Ad}{\text{Ad}}
\newcommand{\hatx}[1]{{[#1]}_{\times}}
\newcommand{\mat}[1]{\mathbf{#1}}
\newcommand{\vect}[1]{\mathbf{#1}}
\newcommand{\eye}{\mat{I}}
\newcommand{\txi}{\boldsymbol{\xi}}
\newcommand{\tdelta}{\boldsymbol{\delta}}
\newcommand{\tomega}{\boldsymbol{\omega}}
\newcommand{\tvec}{\mathbf{t}}

\begin{document}
\title{Exact Higher-Order Derivatives for $SE(3)$ \\
       via Analytical/AD Methods}

\author{Frank~O.~Kuehnel%
\thanks{F. O. Kuehnel is with Sigma Point Labs, Denver, CO, USA
(e-mail: frank@sigmapointlabs.com).}%
\thanks{arXiv preprint, May 2026.}}

\markboth{Preprint --- arXiv 2026}%
{Kuehnel: AD Through Lie Group Priors}

\maketitle

\begin{abstract}
Fast prototyping of new $SE(3)$ estimation objectives remains
awkward in practice. Modern Lie-group frameworks --- GTSAM,
manif, Sophus, SymForce, Ceres --- target first-order workloads
through different code-generation and AD strategies, each
optimized for a particular seam between hand-derived geometry
and generic differentiation. The remaining gap is a compact,
AD-safe path from these first-order primitives to exact
Hessians, observed-information matrices, and higher-order
derivative tensors --- the quantities needed for exact Newton
steps, observed-information covariance estimates, and
covariance correction.
 
This paper presents a hybrid analytical/AD recipe that closes
the gap for $SE(3)$ negative log-likelihoods. The practitioner writes the NLL gradient once, generic over an AD scalar
type, and places the analytical/AD seam at the point-action interface \(\vect y=\mat T\vect x\). Closed-form
Lie-group Jacobians are used up to this interface; AD is applied
only beyond it. The same source is then instantiated with three
scalar types: ordinary floats for gradients, vector-seeded dual
numbers for exact Hessians in a single forward-mode pass, and
nested dual numbers for the higher-order derivative tensors used
in covariance correction. On a representative 6-DoF, 5-landmark
$SE(3)$ NLL, the seeded-Hessian path is approximately $5\times$
faster than finite-differencing the AD gradient on this
benchmark while matching a nested-AD oracle to machine
precision. The implementation adds roughly 70 lines of
analytical-Jacobian code over an AD-only baseline. We also
identify and fix a removable singularity in the standard
$SO(3)/SE(3)$ scalar basis that would otherwise produce
\texttt{NaN}s at the origin under seeded AD, and we audit which
Lie-group derivative tensors require this stabilized basis. The
result is a practical path from rapidly written $SE(3)$
objectives to exact higher-order derivatives, with predictable
runtime and no finite-difference tuning.
\end{abstract}


\section{Introduction}
\label{sec:intro}

The practitioner writing a pose estimator on $SE(3)$ faces a
derivative accounting problem. The NLL for a typical robotics MAP
problem decomposes as
\begin{equation}
-\log p(\mat{T} \mid \mathcal{D}) = 
\sum_i \rho\bigl(\|\mat{L}_i(\pi(\mat{T}\vect{x}_i) - \vect{z}_i)\|^2\bigr)
+ \tfrac{1}{2}\,\txi_{\text{err}}^\top \boldsymbol{\Sigma}^{-1}\txi_{\text{err}},
\label{eq:nll}
\end{equation}
where, for $\mat{T}=(\mat{R},\vect{p})$, 
$\mat{T}\vect{x}:=\mat{R}\vect{x}+\vect{p}$ denotes the standard
$SE(3)$ point action. The first term is the data contribution (sum over
reprojection residuals, with $\mat{L}_i$ the whitening factor satisfying
$\mat{L}_i^\top\mat{L}_i = \boldsymbol{\Omega}_i$ and $\rho$ a robust
kernel); the second is the prior contribution on the manifold
($\txi_{\text{err}} = \Log(\mat{T}^{-1}\mat{T}_{\text{prior}})$ with
$\mat{T}_{\text{prior}}$ denoting the prior mean, distinct from the current
pose $\mat{T}$ being estimated). This form covers the common local building
blocks of robotics state estimation and pose optimization: measurement
residuals, pose priors, covariance estimates, and uncertainty-aware
linearizations.

Two engineering choices dominate practice, each optimized for a
specific regime. The \emph{fully analytical} approach hand-derives
every Jacobian and chains them via the manifold chain rule:
efficient and numerically controlled at the cost of repeating
the derivation work whenever the kernel $\rho$, projection $\pi$,
or residual form changes. The \emph{fully automatic} approach
traces AD through $\Exp$, $\Log$, and the robust kernels: maximally
flexible, at the cost of taping through the trigonometric branches
of $\Exp$ and $\Log$, which carries runtime and numerical
implications near $\theta = 0$. Both choices serve their target
workloads well; the gap this paper addresses is a third regime
where exact higher-order derivatives are needed and neither
choice is a comfortable fit.

The productive middle path is well-known to anyone who has
implemented it: provide closed-form Jacobians of the Lie group
primitives, then use forward-mode AD for the application-specific
cost. The geometric Jacobians act as tangent-space \emph{seeds}
that AD propagates through the user's NLL expression. What is less
standard is choosing this partition so that the same source code
remains valid under seeded and nested AD, producing exact Hessians
and higher-order tensors without finite differences.

\subsection{What Breaks Between First-Order Convenience and Higher-Order Need}
\label{sec:problem}
 
The middle path described above is well understood at first order.
A practitioner can wire up Gauss--Newton or Levenberg--Marquardt
for a new residual on $SE(3)$ in a few hours: hand-roll the
Lie-group Jacobian, let AD propagate through the application-specific
residual, get a first-order Jacobian, hand it to the solver. The
mismatch appears one derivative order higher.
 
Exact Newton steps need a true Hessian, not the
$\mat{J}^\top\mat{J}$ Gauss--Newton approximation.
Observed-information covariance estimates need the same Hessian.
Higher-order tensors, such as the third derivatives that arise in
covariance-correction expansions on $SE(3)$, require yet another
derivative pass. In current practice, each of these either restarts
a hand-derivation chain or routes through finite differences on
the solver's hot path.
 
This paper keeps the middle-path partition, but chooses the
analytical/AD seam so that the partition remains compositional
under seeded and nested AD. The guiding rule is simple: carry the
Lie-group part analytically up to an ordinary vector-valued
interface, then let AD propagate through the application-specific
NLL. For measurement terms, this interface is the point action
$\vect{y}=\mat{T}\vect{x}$; for pose-prior terms, it is the
standard $\Log$ residual together with its closed-form tangent
Jacobian. With this seam placement, the same templated gradient
body yields the gradient, the exact Hessian, and the third-order
tensor by changing only the AD scalar type at the call site ---
provided the analytical Lie-group layer is itself AD-safe. The
bulk of the paper develops this recipe and identifies the
basis-level conditions under which it works.

\subsection{Relation to Existing Frameworks}
\label{sec:related}
 
Modern $SE(3)$ estimation frameworks make different choices about
where the analytical Lie-group machinery ends and the user's
templated code begins. Ceres~\cite{ceres} exposes templated cost
functors and relies on forward-mode AD through user code,
implicitly placing the seam at the user-function boundary --- AD
must therefore tape through any Lie-group calls inside.
GTSAM~\cite{gtsam_expressions} composes closed-form Lie-group
Jacobians through expression graphs, placing the seam at the
factor boundary. manif~\cite{manif} and Sophus~\cite{sophus} ship
analytically stabilized $SO(3)$/$SE(3)$ primitives with
first-order Jacobians, leaving the seam-placement choice to the
user. SymForce~\cite{symforce} generates symbolic Lie-group code
and first-order Jacobians at build time (used, e.g., for the
covariance Jacobians of PX4-ECL/EKF2~\cite{px4ecl}), placing the
seam at the symbolic-spec boundary. All four target first-order
workloads; the seam placement at the point-action interface
$\vect{y} = \mat{T}\vect{x}$ pursued in this paper is a different
trade-off, oriented toward exact Hessians and higher-order tensors
rather than first-order Jacobian assembly.

\section{The Minimal Analytical Core}
\label{sec:core}

The recipe sketched in Section~\ref{sec:problem} requires
closed-form Lie-group primitives up to the point-action
interface $\vect{y} = \mat{T}\vect{x}$. Any of the libraries
named in Section~\ref{sec:related} can supply them. To keep the
discussion below self-contained --- and because we will later
need to refer to specific scalar functions inside these
primitives --- we collect the relevant formulas in this section
in their conventional closed form. The derivations are
standard~\cite{sola2018micro, barfoot2024}; we state them only
to fix notation.
 
A practical convenience is to name the trigonometric quotients
that recur throughout the closed forms.  The literature uses
several naming conventions for these quantities,\footnote{The
closest parallel naming we are aware of is the $\{a,b,c,e\}$
basis tabulated in Jackson's Lie-group identity
tables~\cite{jackson_lietables}: under that convention $a=A$,
$b=B$, and $c=C$, but his $e=(b-2c)/(2a)$ differs from our
$D$, the direct $[\tomega]_\times^2$ coefficient in
$\mat{J}_r^{-1}$.}
so we adopt a local one.  We call the named quotients the
\emph{scalar function basis}, or simply the \emph{scalar basis}.
The term is only an implementation convention: it means the
small set of scalar functions used to assemble the closed forms,
not an algebraically independent basis.

\subsection{Scalar Basis on $SO(3)$}
\label{sec:so3basis}

Let $\theta = \|\tomega\|$, and write $\mat{R}=\exp(\hatx{\tomega})$ and
$\tomega=\Log_{SO(3)}(\mat{R})$ on the standard local branch. The
$SO(3)$ exponential and right Jacobian use the following three scalar
functions~\cite{sola2018micro, barfoot2024}:
\begin{equation}
  A=\frac{\sin\theta}{\theta},\quad
  B=\frac{1-\cos\theta}{\theta^2},\quad
  C=\frac{\theta-\sin\theta}{\theta^3},
\label{eq:ABC}
\end{equation}
with the dependence on $\theta$ suppressed in formulas below. Then
\[
  \mat{R} = \eye + A\hatx{\tomega}
  + B\hatx{\tomega}^2,
  \quad
  \mat{J}_r = \eye - B\hatx{\tomega}
  + C\hatx{\tomega}^2 .
\]
The inverse right Jacobian introduces the remaining $SO(3)$ scalar:
\begin{equation}
  D=
  \frac{1}{\theta^2}\!\left(1 - \frac{\theta}{2}\cot\frac{\theta}{2}\right),
  \quad
  \mat{J}_r^{-1}
  = \eye + \tfrac{1}{2}\hatx{\tomega} + D\hatx{\tomega}^2 .
\label{eq:Jrinv}
\end{equation}

It is useful to regard these scalars as functions of
$s=\theta^2=\tomega^\top\tomega$. They are not algebraically
independent; for instance,
\[
    A+sC=1,\; A^2-2B+sB^2=0, \;
    D=\frac{1}{s}\left(1-\frac{A}{2B}\right).
\]
We nevertheless keep all four names as implementation atoms for the
closed forms and their \(s\)-native AD-safe variants.
Throughout, \(\vect e_m\) denotes the \(m\)-th standard basis
vector in \(\real^3\), \(\mat E_m=[\vect e_m]_\times\), and
\(\mat S_m=\partial[\tomega]_\times^2/\partial\omega_m\).

\subsection{$SE(3)$ Block Structure}
\label{sec:se3block}

We use the rotation-first tangent convention
$\txi = [\tomega^\top, \tvec^\top]^\top \in \real^6$. For
$\mat{T} = (\mat{R}, \vect{p}) \in SE(3)$, the exponential map
is
\[
\Exp(\txi) = \bigl(\Exp_{SO(3)}(\tomega),\,\mat{V}(\tomega)\tvec\bigr),
\quad
\mat{V} = \eye + B\hatx{\tomega} + C\hatx{\tomega}^2,
\]
where $\mat{V}$ is also the left $SO(3)$ Jacobian. With the same
local branch, the logarithm is
\[
\Log(\mat{T}) =
  \begin{bmatrix}
  \Log_{SO(3)}(\mat{R}) \\
  \mat{V}^{-1}\vect{p}
  \end{bmatrix},\quad
\mat{V}^{-1} =
  \eye
  -\tfrac{1}{2}\hatx{\tomega}
  +D\hatx{\tomega}^2 .
\]
With this convention, the $SE(3)$ right Jacobian and its inverse
have lower-triangular block form:
\begin{equation}
\mat{J}_r^{SE(3)} =
  \begin{pmatrix}
  \mat{J}_r & \mat{0} \\
  \mat{Q}_r & \mat{J}_r
  \end{pmatrix},\quad
\bigl(\mat{J}_r^{SE(3)}\bigr)^{-1} =
  \begin{pmatrix}
  \mat{J}_r^{-1} & \mat{0} \\
  \widetilde{\mat{Q}}_r & \mat{J}_r^{-1}
  \end{pmatrix},
\label{eq:se3jacinv}
\end{equation}
where $\mat{Q}_r = -\mat{J}_r\widetilde{\mat{Q}}_r\mat{J}_r$.

The remainder of the paper is written on the inverse side.
This is the side needed by the pose-prior seam of
Section~\ref{sec:problem}: differentiating the $\Log$ residual
pulls in $\mat{J}_r^{-1}$ on $SO(3)$ and $\widetilde{\mat{Q}}_r$
on $SE(3)$. Forward-side quantities such as $\mat{Q}_r$ and
$\partial\mat{J}_r/\partial\omega_m$ are obtained by the block
identities above rather than written as separate closed forms.

The inverse coupling block has the compact hat-matrix form
with $\mat H \equiv \hatx{\tomega}\hatx{\tvec}
+ \hatx{\tvec}\hatx{\tomega}$
\begin{equation}
\widetilde{\mat{Q}}_r(\tomega, \tvec) =
  \tfrac{1}{2}\hatx{\tvec}
  + D\mat H + \alpha\,\hatx{\tomega}^2, \quad
  \alpha = (\tomega^\top\tvec)\,\bar{\beta}
\label{eq:Qr_compact}
\end{equation}
where the additional $SE(3)$ fused scalar is
\begin{equation}
\bar{\beta}(s) \equiv \frac{\beta(s)}{s},
\qquad
\beta(s) = \frac{C(s)}{2B(s)} - 2D(s).
\label{eq:beta_bar}
\end{equation}
The implementation must provide $\bar\beta$ directly as
an $s$-native scalar, using for example
\[
  \bar{\beta}(s) =
  \tfrac{1}{360} + \tfrac{s}{7560} + \tfrac{s^2}{201600} + \cdots .
\]
Section~\ref{sec:pattern} explains how this fused-scalar rule is used throughout the derivative tensors.

A small but useful exact identity holds:
\begin{equation}
\bar\beta(s) \equiv \widetilde D(s)
= 2\,\frac{dD}{ds},
\label{eq:beta_D_identity}
\end{equation}
with both sides extended to $s = 0$ by their Taylor series.

At small angles, $D \to 1/12$ and $\bar{\beta} \to 1/360$, so
\[
\widetilde{\mat{Q}}_r \approx
\tfrac{1}{2}\hatx{\tvec}
+ \tfrac{1}{12}\mat H
+ \tfrac{1}{360}(\tomega^\top\tvec)\hatx{\tomega}^2 .
\]
The AD analysis in Section~\ref{sec:traps} depends only on the
scalar content of this expression, namely $D(s)$,
$\bar{\beta}(s)$, and their derivatives.

\subsection{Composition, Adjoint, and Point Action}
\label{sec:adjoint_action}

For composed poses $\mat{T}_{12} = \mat{T}_1\cdot\mat{T}_2$,
let $\txi_i \in \real^6$ denote a right perturbation of
$\mat{T}_i$, defined by $\mat{T}_i \mapsto \mat{T}_i \Exp(\txi_i)$.
The identity
$\Exp(\txi_1)\mat{T}_2 = \mat{T}_2 \Exp(\Ad_{\mat{T}_2^{-1}}\txi_1)$
shows that, to first order, a right perturbation of
$\mat{T}_{12}$ is built from the input perturbations as
\begin{equation}
\delta\txi_{12} \;=\; \Ad_{\mat{T}_2^{-1}}\txi_1 \;+\; \txi_2 ,
\label{eq:compjac}
\end{equation}
i.e.\ $\txi_1$ enters via adjoint conjugation and $\txi_2$
enters with the identity. When the composed pose is
subsequently mapped through $\Log$, the appropriate inverse
right-Jacobian factor on the output coordinate appears in
the chain rule; we make that factor explicit in the prior
gradient~\eqref{eq:prior_gradient} below.

The adjoint representation $\Ad_{\mat{T}}$ is purely
algebraic---no trigonometry at all:
\begin{equation}
\Ad_{\mat{T}} =
  \begin{pmatrix} \mat{R} & \mat{0} \\ \hatx{\vect{p}}\mat{R} & \mat{R} \end{pmatrix}.
\label{eq:adjoint}
\end{equation}

For a point $\vect{x}\in\real^3$, we use the standard point action
$\vect{y} = \mat{T}\vect{x} := \mat{R}\vect{x}+\vect{p}$.
Under a right perturbation $\mat{T}(\tdelta)=\mat{T}\Exp(\tdelta)$,
the familiar point-action Jacobian is
\begin{equation}
\mat{J}_{\rm act}(\mat{T},\vect{x})
\equiv
\left.
\frac{\partial(\mat{T}\Exp(\tdelta)\vect{x})}{\partial\tdelta}
\right|_{\tdelta=0}
=
  \begin{pmatrix}
  -\mat{R}[\vect{x}]_\times & \mat{R}
  \end{pmatrix},
\label{eq:Jact}
\end{equation}
where the columns follow the rotation-first convention
$\tdelta=[\delta\tomega^\top,\delta\tvec^\top]^\top$.

The core primitives collected above --- $\Exp$, $\Log$,
$\mat{J}_r^{SE(3)}$ and its inverse, $\Ad$, and
$\mat{J}_{\rm act}$ --- cover the geometric derivatives that
appear in the $SE(3)$ NLL considered here. Everything downstream
is flat-space calculus handled by AD.

\section{The Analytical/AD Seam}
\label{sec:seam}

The analytical core of Section~\ref{sec:core} and the user-facing
implementation of Section~\ref{sec:pattern} meet at a single
seam: an analytical Lie-group map produces a vector interface,
after which application-specific arithmetic takes over. Both
terms of the NLL in~\eqref{eq:nll} share this structure. An
analytical map produces an ordinary vector,
\begin{equation}
\vect{u} = \phi(\mat{T}),
\label{eq:interface}
\end{equation}
and the application-specific objective is a scalar function
$r(\vect{u})$ of that vector. At a right perturbation
$\mat{T} = \bar{\mat{T}}\,\Exp(\tdelta)$, the gradient factors as
\begin{equation}
\nabla_{\tdelta} r =
\underbrace{
\left.\!\left(\frac{\partial \vect{u}}{\partial \tdelta}\right)^{\!\top}\!
\right|_{\tdelta=\mat{0}}
}_{\text{analytical Lie-group side}}
\;
\underbrace{\nabla_{\vect{u}} r}_{\text{AD/user-code side}}.
\label{eq:gen_seam}
\end{equation}
Equation~\eqref{eq:gen_seam} is the analytical/AD seam. The
Jacobian on the left is supplied by the closed-form Lie-group
machinery of Section~\ref{sec:core}. The vector on the right is
obtained by differentiating the below-seam Euclidean residual,
either by ordinary AD through templated arithmetic or by an explicit flat-space chain rule.
The two terms of~\eqref{eq:nll} arise as two choices of the
interface variable $\vect{u}$:
\begin{equation}
\renewcommand{\arraystretch}{1.3}
\begin{array}{c|c|c}
\text{term} & \text{interface } \vect{u} & \partial \vect{u}/\partial \tdelta \\
\hline
\text{data} & \vect{y}_i = \mat{T}\vect{x}_i &
\begin{pmatrix} -\mat{R}\hatx{\vect{x}_i} & \mat{R} \end{pmatrix} \\
\text{prior} & \txi_{\text{err}} = \Log(\mat{T}^{-1}\mat{T}_{\text{prior}}) &
-\,\bigl(\mat{J}_r^{SE(3)}\bigr)^{-1}\!\Ad_{\mat{G}^{-1}}
\end{array}
\label{tab:seam}
\end{equation}
The remainder of this section works through both rows.

\subsection{Data Term: Seam at the Point Action}
\label{sec:seam_data}

For each measurement contribution, the Lie-group side stops at the
point-action interface $\vect y_i = \mat T\vect x_i$ .
Let
\[
    \vect q_i \equiv \nabla_{\vect y_i} r_i
\]
denote the flat-space gradient of the measurement contribution with
respect to this interface variable. The projection, whitening, robust
kernel, and sensor-specific residual algebra are all contained in the
computation of \(\vect q_i\). They are not part of the reusable
Lie-group layer.

The lift back to pose perturbation coordinates is therefore
\begin{equation}
    \nabla_{\tdelta} r_i
    =
    \mat J_{\rm act}(\mat T,\vect x_i)^\top \vect q_i .
\label{eq:data_lift}
\end{equation}
Changing the residual model changes \(\vect q_i\), but not the
Lie-group lift.

\subsection{Prior Term: Seam at the $\Log$ Residual}
\label{sec:seam_prior}

For the pose prior, the below-seam scalar is the quadratic
\begin{equation}
r_{\text{prior}}(\txi_{\text{err}}) =
\tfrac{1}{2}\,\txi_{\text{err}}^\top \boldsymbol{\Sigma}^{-1}\txi_{\text{err}},
\;
\nabla_{\txi_{\text{err}}} r_{\text{prior}} =
\boldsymbol{\Sigma}^{-1}\txi_{\text{err}}.
\label{eq:prior_scalar}
\end{equation}
The analytical side of~\eqref{eq:gen_seam} is the tangent
Jacobian of the $\Log$ residual; with
$\mat{G} = \bar{\mat{T}}^{-1}\mat{T}_{\text{prior}}$, a standard
right-perturbation
calculation~\cite{sola2018micro, barfoot2024} gives
\begin{equation}
\nabla_{\tdelta} r_{\text{prior}} =
-\,\Ad_{\mat{G}^{-1}}^\top
\bigl(\mat{J}_r^{SE(3)}(\txi_{\text{err}})^{-1}\bigr)^{\!\top}\!
\boldsymbol{\Sigma}^{-1}\txi_{\text{err}}.
\label{eq:prior_gradient}
\end{equation}
The inverse right Jacobian comes from differentiating $\Log$;
the adjoint propagates the right perturbation through inversion
and composition. At first order there is no application-specific
AD work for the prior term: the below-seam function is
quadratic, and the above-seam factors are closed-form Lie-group
quantities. When the same gradient body is recompiled with dual
or nested-dual scalars to obtain the Hessian or third-order
tensor, however, those closed-form factors are themselves
evaluated in the AD scalar type \texttt{S}; higher derivatives
then differentiate through the analytical formulas
$\mat{J}_r^{-1}$ and $\widetilde{\mat{Q}}_r$ rather than through
finite differences. Whether this propagation is safe depends on
how those formulas are written --- a question we take up in
Section~\ref{sec:pattern}.

\subsection{One Source, Three Derivative Orders}
\label{sec:three_orders}

The two choices of interface variable,
\[
    \vect y_i = \mat T\vect x_i
    \qquad\text{and}\qquad
    \txi_{\rm err} = \Log(\mat T^{-1}\mat T_{\rm prior}),
\]
allow the NLL gradient to be written as one scalar-generic body
\(g(\tdelta)\), with signature schematically
\[
    \texttt{nll\_grad<S>(delta: Vec6<S>) -> Vec6<S>}.
\]
Changing only the scalar type \texttt{S} yields three derivative
artifacts:
\begin{itemize}
\item \texttt{S = f64}: ordinary evaluation returns the gradient
      \(g(\mat 0)\).
\item \texttt{S = Dual6}: one forward-mode pass returns the gradient
      together with the exact \(6\times6\) Hessian
      \(\partial g/\partial\tdelta|_{\tdelta=\mat 0}\).
\item \texttt{S = NestedDual6}: a further forward-mode pass returns the
      third-order tensor
      \(\partial^2 g/\partial\tdelta^2|_{\tdelta=\mat 0}\), which appears
      in higher-order covariance-correction expansions.
\end{itemize}
The runtime comparison is reported in Section~\ref{sec:validation}. The
construction works only if the analytical side of each seam ---
\(\mat J_{\rm act}\), \(\mat J_r^{-1}\), \(\widetilde{\mat Q}_r\), and
the derivative tensors of the last two --- is itself safe to evaluate
with the same scalar type \texttt{S}. The next section addresses how to
make that the case.

\section{The User-Facing Pattern}
\label{sec:pattern}

The seam established in Section~\ref{sec:seam} is realized as the
scalar-generic gradient body described in
Section~\ref{sec:three_orders}. The key point is that the differentiated
object is the assembled gradient \(g(\tdelta)\), not a scalar objective
traced naively through all of \(\Exp\), \(\Log\), projection, and
robust-loss code. The Lie-group side is kept explicit; the below-seam
residual calculus is ordinary flat-space arithmetic.

\subsection{The Algorithm and Its Requirements}
\label{sec:listing}

Algorithm~\ref{alg:nll_grad} shows the computational structure. The AD
decision variable is the local right perturbation \(\tdelta\) at a fixed
linearization pose \(\bar{\mat T}\), so that
\(\mat T(\tdelta)=\bar{\mat T}\Exp(\tdelta)\). The solver refreshes
\(\bar{\mat T}\) between Newton steps; AD only sees \(\tdelta\). There
is no separate AD step in the algorithm: the derivative order is
determined solely by the scalar type \texttt{S} used to evaluate the
same body.

\begin{algorithm}[t]
\caption{AD-generic analytical NLL gradient. The algorithm shows the
seam structure; the scalar type \texttt{S} supplies the derivative
payload.}
\label{alg:nll_grad}
\begin{algorithmic}[1]
\Require scalar type \texttt{S}; perturbation
\(\tdelta\in\texttt{Vec6<S>}\); base pose \(\bar{\mat T}\); prior
\((\mat T_{\rm prior},\boldsymbol{\Sigma}^{-1})\); observations
\(\mathcal O\)
\Ensure gradient \(g(\tdelta)\in\texttt{Vec6<S>}\)

\State \(\mat T \gets \bar{\mat T}\Exp(\tdelta)\)
\State \(\vect g_{\rm body} \gets \vect 0\)

\Statex
\Statex \textit{Prior seam: \(\txi=\Log(\mat T^{-1}\mat T_{\rm prior})\)}
\State \(\mat G \gets \mat T^{-1}\mat T_{\rm prior}\)
\State \(\txi \gets \Log(\mat G)\)
\State \(\mat J_{\rm prior}
       \gets
       -\Ad_{\mat G^{-1}}^\top
       \bigl(\mat J_r^{SE(3)}(\txi)^{-1}\bigr)^\top\)
\State \(\vect g_{\rm body}
       \gets
       \vect g_{\rm body}
       +
       \mat J_{\rm prior}\boldsymbol{\Sigma}^{-1}\txi\)

\Statex
\Statex \textit{Measurement seams}
\For{each observation \(o\in\mathcal O\)}
    \State \((\vect u,\mat J_{\rm geom})
       \gets
       o.\mathrm{geometricInterface}(\mat T)\)
    \State \(\vect q
       \gets
       o.\mathrm{gradientBelowSeam}(\vect u)\)
    \State \(\vect g_{\rm body}
       \gets
       \vect g_{\rm body}
       +
       \mat J_{\rm geom}^{\top}\vect q\)
\EndFor

\Statex
\Statex \textit{Parameterization}
\State \Return
       \(\bigl(\mat J_r^{SE(3)}(\tdelta)\bigr)^\top
       \vect g_{\rm body}\)
\end{algorithmic}
\end{algorithm}

Here \(o.\mathrm{geometricInterface}\) returns the vector interface
\(\vect u=\phi_o(\mat T)\) and its analytical Lie-group Jacobian
\(\mat J_{\rm geom}\). For a point-action residual,
\(\vect u=\mat T\vect x_i\) and
\(\mat J_{\rm geom}=\mat J_{\rm act}(\mat T,\vect x_i)\). The call
\(o.\mathrm{gradientBelowSeam}(\vect u)\) returns the Euclidean gradient
\(\vect q=\nabla_{\vect u}r_o\). The reusable Lie-group layer only sees
\(\vect u\), \(\mat J_{\rm geom}\), and \(\vect q\).

The accumulated \(\vect g_{\rm body}\) is a local right-perturbation
gradient at \(\mat T(\tdelta)\). The returned gradient, however, is with
respect to the coordinates \(\tdelta\) in
\(\mat T(\tdelta)=\bar{\mat T}\Exp(\tdelta)\), hence the final
parameterization factor
\[
    g(\tdelta)
    =
    \bigl(\mat J_r^{SE(3)}(\tdelta)\bigr)^\top
    \vect g_{\rm body}.
\]
This factor must remain inside the AD-generic function: when seeded AD
differentiates the returned gradient, it also differentiates
\(\mat J_r^{SE(3)}(\tdelta)^\top\), producing the parameterization terms
required for the exact Hessian.

\subsection{Genericity or Collapse}
\label{sec:genericity}

A single \texttt{f64}-specialized primitive breaks the construction. If,
for example, \texttt{Exp} accepts only \texttt{Vec6<f64>} rather than
\texttt{Vec6<S>}, then dual scalars cannot pass through the call. Worse,
an implicit cast to \texttt{f64} may silently erase derivative
information. The remedy is mechanical but strict: every primitive used
by the gradient body must be generic over the scalar type \texttt{S},
including \texttt{Exp}, \texttt{Log}, \texttt{Jr}, \texttt{JrInv},
\texttt{Ad}, the hat operator, and the point action.

This requirement is structural, not numerical. No special input triggers
the failure; one non-generic function anywhere on the path from
\(\tdelta\) to \(g(\tdelta)\) destroys the single-source property.
The scalar type \texttt{S} must support ordinary arithmetic, numeric
constants, and the elementary functions used on the regular branch. The
additional small-angle requirements are part of the scalar-basis audit
below.

\subsection{Hazards in the Scalar Basis and Their Fixes}
\label{sec:traps}

When seeded-mode AD differentiates the analytical gradient, two hazards
appear in the closed-form Lie-group layer: removable singular factor
pairs and insufficient small-angle Taylor degree. Both come from
conventional \(\theta\)-based formulas, not from the smooth maps
themselves.

\paragraph{Singular factor pairs}
Derivative tensors such as
\(\partial\mat J_r^{-1}/\partial\omega_m\) and
\(\partial\widetilde{\mat Q}_r/\partial\omega_m\) contain scalar factors
of the form
\[
    \frac{D'(\theta)\,\omega_m}{\theta}.
\]
The product is smooth, but the factors are not:
\(D'(\theta)=\theta/360+O(\theta^3)\), while
\(\omega_m/\theta\) is undefined at \(\tomega=\mat 0\). AD evaluates the
factors separately, so the origin may produce \(0\cdot(0/0)\) and inject
a NaN into the seeded Hessian.

The fix is to fuse the singular pair before AD sees it. Work in
\(s=\tomega^\top\tomega\) and implement
\begin{equation}
    \widetilde D(s)
    \equiv
    \frac{D'(\theta)}{\theta}
    =
    2\frac{dD}{ds}
    =
    \frac{1}{360}
    +
    \frac{s}{7560}
    +
    \frac{s^2}{201600}
    + \cdots .
\label{eq:Dtilde}
\end{equation}
Then every occurrence of \(D'(\theta)\omega_m/\theta\) is written as
\(\widetilde D(s)\omega_m\). No division by \(\theta\) and no
\(\sqrt{s}\) occur on the singular path.

For example, the implemented \(SO(3)\) inverse-right-Jacobian derivative
is
\begin{equation}
\frac{\partial \mat J_r^{-1}}{\partial \omega_m}
=
\tfrac{1}{2}\mat E_m
+
\widetilde D(s)\,\omega_m[\tomega]_\times^2
+
D(s)\mat S_m .
\label{eq:Jrinv_deriv}
\end{equation}

The inverse \(SE(3)\) coupling-block derivative follows from the same fused rule. Define
$\mat H_m \equiv \mat E_m\hatx{\tvec}+\hatx{\tvec}\mat E_m$, then
writing
\[
    \alpha_m
    \equiv
    \frac{\partial\alpha}{\partial\omega_m}
    =
    t_m\bar\beta(s)
    +
    2\omega_m(\tomega^\top\tvec)\bar\beta_s(s),
    \quad
    \bar\beta_s(s)=\frac{d\bar\beta}{ds},
\]
the \(s\)-native derivative is
\begin{equation}
\frac{\partial\widetilde{\mat Q}_r}{\partial\omega_m}
=
\widetilde D(s)\omega_m\mat H
+
D(s)\mat H_m
+
\alpha_m[\tomega]_\times^2
+
\alpha\mat S_m .
\label{eq:Qtilde_deriv}
\end{equation}
No division by \(\theta\) appears. The equivalent \(\theta\)-quotient
formulas are useful for derivation, but not for seeded evaluation.

\paragraph{Polynomial degree on the small-angle branch}
Small-angle branches are Taylor polynomials in \(s\). A degree-\(n\)
polynomial has zero \((n+1)\)-st derivative with respect to \(s\), so a
branch intended for AD depth \(k\) must retain degree at least \(k\). For
the Hessian path, degree two is the minimum; the implementation uses
degree four as padding for higher-order tensors.

\subsection{The Recipe in Five Steps}
\label{sec:recipe}

The implementation audit is:

\begin{enumerate}
\item Make every primitive generic over the scalar type, including
algebraic helpers such as \(\Ad\), the hat operator, and the point action.
\item Parameterize scalar functions by \(s=\theta^2=\tomega^\top\tomega\),
not by \(\theta\), on the small-angle branch.
\item Extend each small-\(s\) Taylor branch to degree \(\geq k\) for
target AD depth \(k\).
\item Replace every removable quotient \(f(\theta)/\theta^k\) in the
derivative tensors by a fused \(s\)-native scalar, such as
\(\widetilde D(s)=D'(\theta)/\theta\) or
\(\bar\beta(s)=\beta(s)/s\).
\item Verify the formulas once away from the origin and once in the
small-angle basin, e.g. with \(s<10^{-8}\).
\end{enumerate}

After this audit, the analytical gradient can be safely instantiated with
dual scalar types. The next section measures the resulting Hessian path.

\section{Validation: AD Through NLL at Depth 2}
\label{sec:validation}

We validate the recipe of Section~\ref{sec:pattern} at derivative
depth two, i.e. for the Hessian, on a representative 6-DoF
\(SE(3)\) NLL with a concentrated Gaussian prior and five PnP
landmark terms. The data term uses the smooth pseudo-Huber kernel
\begin{equation}
\rho(s)=\kappa^2\bigl(\sqrt{1+s/\kappa^2}-1\bigr),
\quad
s=\|\mat{L}_i(\pi(\mat{T}\vect{x}_i)-\vect{z}_i)\|^2 ,
\label{eq:pseudo_huber}
\end{equation}
so the comparison is not affected by nonsmooth robust-loss corners.
Table~\ref{tab:validation} is not a portable microbenchmark:
absolute runtimes depend on hardware, compiler, scalar type, and AD
implementation details. We therefore report only relative speed
against the FD-of-AD-gradient baseline.
LOC is reported only to indicate the size of the handwritten analytical
surface area; it is not intended as a language-independent complexity
measure.

\begin{table*}[t]
\caption{Hessian of a 6-DoF, 5-landmark \(SE(3)\) NLL computed by
several derivative paths. Relative error is measured against the
nested-AD fused-basis oracle in Row~5. Speed is relative to Row~2;
values larger than one are faster.}
\label{tab:validation}
\centering
\begin{tabular}{llrll}
\toprule
\# & Method & LOC & Rel.\ err.\ vs.\ oracle & Speed vs.\ Row~2 \\
\midrule
1 & FD of value (no AD)
  & 43  & \(6.65\times 10^{-3}\) & \(0.95\times\) \\

2 & FD of AD-gradient (baseline)
  & 30  & \(9.18\times 10^{-7}\) & \(1.00\times\) \\

3 & FD of analytical gradient (fused basis)
  & 104 & \(9.18\times 10^{-7}\) & \(2.82\times\) \\

4 & Nested AD, na\"ive basis
  & 118 & \(\sim 10^{-15}\) & \(0.83\times\) \\

5 & Nested AD, fused basis (oracle)
  & 30  & 0 (reference) & \(0.83\times\) \\

6 & Seeded AD of analytical gradient, na\"ive basis
  & \((102)^\dagger\) & NaN & --- \\

7 & \(\star\) Seeded AD of analytical gradient, fused basis (recipe)
  & 102 & \(\mathbf{2.41\times 10^{-16}}\) & \(\mathbf{4.83\times}\) \\

8 & Auto FoR (reverse-mode tape, no analytical gradient)
  & 30  & \(2.75\times 10^{-16}\) & \(0.79\times\) \\
\bottomrule
\end{tabular}

\vspace{0.3em}
\footnotesize
LOC counts user-written analytical lines, excluding the AD scalar type
implementation itself.
\(\dagger\) Row~6 mirrors Row~7 structurally, but the unfused
\(\theta\)-based factor produces \(0/0\) at the origin before useful
seeded differentiation can proceed.
\end{table*}

Row~7 is the recipe advocated in this paper: seeded forward-mode AD is
applied to the assembled analytical gradient, with the fused
\(s=\tomega^\top\tomega\) scalar basis. The correctness check is
Row~7 against Row~5: the proposed path agrees with the nested-AD oracle
to machine precision. The failure-mode check is Row~6 against Row~7:
the same seeded strategy fails with the conventional \(\theta\)-based
scalar form and succeeds once the removable singular factor pairs are
fused. This is the central validation point. AD differentiates the
computation graph it is given; the Lie-group layer must therefore be
written in a form that is differentiable to the required order.

Rows~1--3 show the finite-difference alternatives. Finite differences
of the value are too inaccurate for this Hessian, while finite
differences of either gradient reach only the expected FD accuracy.
These baselines are useful but not the target use case: the recipe is
meant to avoid finite-difference step-size tuning on the solver path.

Rows~4, 5, and 8 show that other AD strategies can also produce the
correct Hessian at depth two. In particular, Row~8 confirms that
forward-over-reverse is a valid conventional Hessian strategy when a
reverse-mode tape is available: it differentiates a reverse-mode
gradient of the scalar NLL rather than the analytical gradient body.
The comparison should not be read as a fundamental statement about
reverse mode: its cost is highly engineering-sensitive, depending on
tape layout, allocation, reference counting, scratch reuse, and cache
locality. The narrower point is that Row~7 obtains the same Hessian
accuracy without finite differences or reverse-mode infrastructure,
while keeping the Lie-group layer explicit and auditable for
higher-order seeded derivatives.

\paragraph{Formula-level cross-check.}
As an independent check on the analytical core, we also compared the
closed-form inverse right Jacobian against the Lie-algebra matrix
function
\begin{equation}
\bigl(\mat{J}_r^{SE(3)}(\txi)\bigr)^{-1}
= \frac{\mathrm{ad}_{\txi}}{\mat{I} - \exp(-\mathrm{ad}_{\txi})}
= \mat{I} + \tfrac{1}{2}\mathrm{ad}_{\txi}
+ \tfrac{1}{12}\mathrm{ad}_{\txi}^{2} + \cdots,
\label{eq:adseries}
\end{equation}
evaluated by a Bernoulli-series approximation. Across the tested
$(\tomega,\tvec)$ regimes, the result agreed with the block
decomposition~\eqref{eq:se3jacinv} and with the compact
$\widetilde{\mat{Q}}_r$ form~\eqref{eq:Qr_compact} to the accuracy
expected from the truncated series.

Thus, with the analytical core of Section~\ref{sec:core} and the
basis fixes of Section~\ref{sec:pattern}, the same NLL source yields a
self-consistent gradient and Hessian. The timing numbers in
Table~\ref{tab:validation} should be read only as evidence that the
recipe is practically competitive in this implementation, not as a
portable claim about absolute runtime.
A companion implementation of the benchmark and scalar-basis checks is
available in~\cite{se3_ad_recipes}.

\section{Discussion}
\label{sec:discussion}

The point of this paper is not that automatic differentiation is
insufficient. Forward mode, reverse mode, symbolic code generation, and
mixed strategies such as forward-over-reverse are all effective tools,
and modern robotics libraries use them well. The point is more specific:
for Lie-group estimation, AD differentiates the program that is written,
not the smooth geometric map one has in mind. The choice of coordinates,
the placement of the analytical/AD seam, and the scalar form of the
closed-form primitives still matter.

The recipe developed here occupies this middle ground. The reusable
$SE(3)$ geometry is kept analytical up to the point-action and Log-residual
interfaces, while the application-specific objective remains ordinary
templated code. This is not intended to replace first-order
Gauss--Newton or Levenberg--Marquardt workflows in Ceres, GTSAM, SymForce,
Sophus, or related libraries. For those use cases, existing tools are
usually sufficient. The intended regime is narrower: exact Hessians,
observed-information matrices, and higher-order uncertainty corrections,
where finite differences are noisy or expensive and fully nested AD
through the Lie-group layer can become costly or numerically fragile.

The validation reflects this distinction. On the benchmark in
Table~\ref{tab:validation}, the fused seeded-AD path matches the
nested-AD oracle to machine precision and is roughly $5\times$ faster
than the FD-of-AD-gradient fallback. The speedup itself is not the main
claim; it is problem-dependent and should be remeasured in other
settings. The more important result is structural: the naive
$\theta$-based analytical formulas fail under seeded AD, while the
$s=\omega^\top\omega$ fused scalar basis makes the same analytical
gradient safe to differentiate. This is the concrete sense in which
``just use AD'' is not a complete answer for higher-order Lie-group
derivatives.

Derivative depth strengthens the case for the recipe. At depth two,
nested AD remains a reasonable oracle and forward-over-reverse is a
natural Hessian strategy when a reverse-mode tape is available. At depth
three and beyond, the tensors needed for covariance correction are no
longer produced by a single Hessian pass, and nested scalar payloads grow
rapidly. Seeded evaluation of an analytical gradient keeps the
differentiation path explicit and fixed, provided the small-angle
branches have sufficient polynomial degree and the removable
singularities have been eliminated before AD sees them.

The cost is a small analytical core and a corresponding maintenance
contract. In the implementation measured here, the analytical path adds
about $70$ lines over the AD-only baseline. Those lines must remain
generic over the scalar type, use $s$-native small-angle branches, and
preserve the fused scalar substitutions used by the derivative tensors.
For projects that only need occasional first-order derivatives, this is
unnecessary overhead. For projects that repeatedly need exact Hessians
or higher-order tensors on $SE(3)$, the audit is a one-time cost that
removes finite-difference tuning and avoids differentiating through
ill-conditioned algebraic forms.

Finally, the scalar-basis fixes address the origin and the
small-angle basin; they do not solve every numerical issue on
$SE(3)$. In particular, conditioning near the rotation-chart
boundary ($\theta \approx \pi$) remains a separate problem,
addressed by the usual remedies of chart switching or
regularization. The contribution here is therefore deliberately
limited: it is an AD-safe higher-order
derivative pattern for the common local regime of $SE(3)$ optimization,
not a universal replacement for existing solver architectures.
 
\section*{Acknowledgments}

The author thanks André Jalobeanu for very useful discussions and
early feedback that helped clarify the failure modes diagnosed
in this paper.

\bibliographystyle{IEEEtran}
\bibliography{references}

\end{document}